\documentclass{elsarticle}
\usepackage[utf8]{inputenc}
\usepackage{tabularx}
\usepackage{longtable}
\usepackage{booktabs}
\usepackage[numbers]{natbib}
\usepackage{graphicx}
\title{Review of Computational Epigraphy}
\author{Vishal K}
\date{\today}

\begin{document}

	\begin{abstract}

	Computational Epigraphy refers to the process of extracting text from stone inscription, transliteration, interpretation, and attribution with the aid of computational methods. Traditional epigraphy methods are time consuming, and tend to damage the stone inscriptions while extracting text. Additionally, interpretation and attribution are subjective and can vary between different epigraphers. However, using modern computation methods can not only be used to extract text, but also interpret and attribute the text in a robust way. We survey and document the existing computational methods that aid in the above-mentioned tasks in epigraphy.
 \end{abstract}
 \maketitle
	\section{Introduction}
	\subsection{Epigraphs}
	Stone inscriptions are important artifacts in the field of archaeology. Although several cultures follow different methods as primary forms of writing, for example, palm leaf manuscripts by Dravidians, papyrus manuscripts by Egyptians, and Animal Hide manuscripts by several European civilizations, stone inscriptions remained a robust secondary form of writing across all the civilizations that practiced writing. This is mainly due to the robustness associated with the medium, as it is impossible to manipulate, change the stone inscriptions and very difficult to mutilate them. Therefore, several historically important documents such as treaties, grants, and tombstones are engraved in stones. 
	\subsection{Interpretation of Epigraphs}
	Traditional epigraphy practice involves the use of inks to take the imprint of the inscription. This process has a high probability of mutilating the inscription. After that, an expert epigrapher will try to interpret the characters in the inscription.
	\subsection{Challenges in Interpreting Epigraphs}
	Traditional epigraphy practice involves the use of inks to take the imprint of the inscription. This process has a high probability of mutilating the inscription. After that, an expert epigrapher will try to interpret the characters in the inscription.
	\subsection{Time and Place of origin}
	Adding on to that, though written within the same script, inscriptions can vary according to the location and time of origin, since people who lived in different eras and different parts of the same territory will practice different dialects of a script. Therefore, attributing the inscription with their time and place of origin is also important while interpreting the inscription. But it is challenging because several stone inscriptions have been moved from their location of origin in due course of history. 
	\subsection{Mutilation}
	Similarly, mutilation of inscriptions also proved to be a serious factor while interpreting the inscriptions. As we have seen above, misinterpreting a few characters might change the meaning of the inscription. But stone inscriptions are supposed to undergo mutilation due to wear and tear and due to humans (invading armies, accidents, etc.)
	\subsection{Computational Epigraphy}
	The recent rise of computational imaging and computational linguistics, which are in turn caused by the rise of machine learning, has begun to be used for epigraphy. Hence, a new field of computational epigraphy is at its infancy. These technologies don’t aim to replace the epigraphers. But they are aimed to assist them in their work. We have divided computational epigraphy into two domains. One is to transliterate the characters present in the inscription. This involves taking the image of the stone inscription, pre-processing, binarizing, denoising, segmenting individual characters and recognizing those characters. The second part focuses on assigning attributes to the transliterated text. These attributes can be time and place of origin, identities of the named entities in the text, finding the missing text, predicting the sequence of multiple texts, etc.
	
	\section{Literature Review}
	We have compiled a list of papers in the field of computational epigraphy. This does not include papers from adjacent field of Handwritten character recognition, because, even though they’re similar in some ways, stone inscriptions provide unique challenges due to the medium. In addition, we have collected the research regardless of the language and script being used. 
	The following figure shows the distribution of languages/scripts in which computational epigraphy is performed in our literature. 
	\subsection{Transliteration}
	Transliteration is the process of identifying individual characters in the stone epigraph. It may involve mapping the characters from the old script to the modern script of the same language, if that language still exists. 
	
	For example, while transliterating Tamil epigraphs, the characters are mapped to modern Tamil characters, as the phonetics of the characters and grammar of the language remained the same. However, it is not always possible, as some languages became extinct and do not have modern form. For example, Egyptian hieroglyphs cannot be mapped to modern characters. 
	
	Transliteration in Computational Epigraphy often involves, imaging the inscription, processing the image, segmentation at line, word and character level and identifying individual characters. As mentioned earlier several interpretations of the same character can exist; therefore, it is desirable to have explanations for each interpretation. 
	\subsubsection{Cryptography}
\begin{figure}
    \centering
    \includegraphics[scale=0.3]{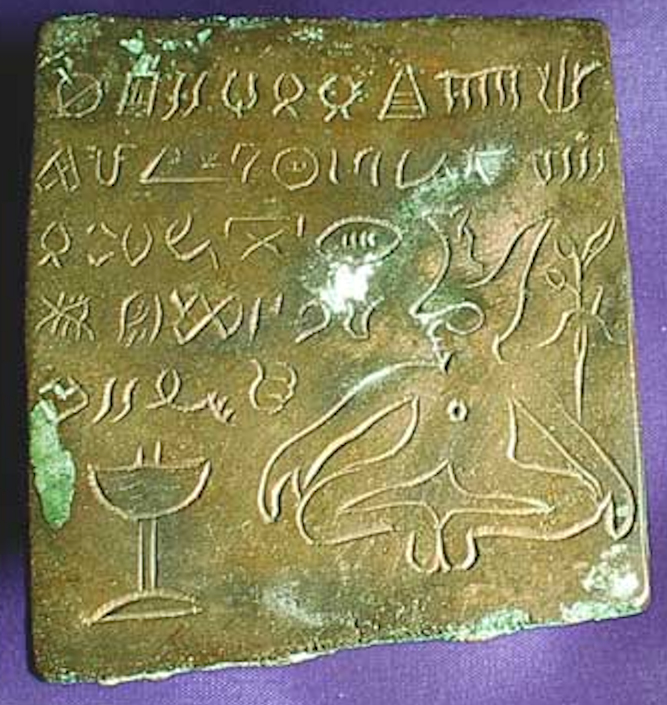}
    \caption{Copper plate with Indus Valley Script}
    \label{fig:indus}
\end{figure}

	Transliteration of  Indus script poses an unique challenge. Because, unlike most other scripts in this paper, the Indus script has not been deciphered yet. There are no multilingual epigraphs identified belonging to the Indus Valley Civilization. Also, the scripts evolved from Indus script like Brahmi are widely changed from the original Indus script. Therefore, it remains undecipherable till the date. Figure \ref{fig:indus} shows an inscription discovered at one of the excavation sites, which has 34 Indus Valley Script characters, which is longest discovered till the date
	
	However, several statistical studies have been made to identify patterns in Indus script, which we will see in later section. In this paper, \cite{cryptographic2010chadha} , the authors try to decipher Indus Script using Cryptography. 
	
	Cryptography was a field emerged during World War II. It is discovered as a tool to decipher the secret messages of Nazi Germany from Enigma Machine \cite{CryptanalysisEnigma2022} by Alan Turing and others. 
	
	Here the author considers an Indus script similar to Enigma code and try to decipher the key using crypt analysis. They consider each symbol to be a mathematical code and try to decipher them in a known language. Though they made some progress, eventually it failed as Indus script proved to be too complex to be solved using symbolic mathematics.

	\subsubsection{Cuneiform Transliteration}
	 Another script that is challenging to transliterate is Cuneiform. It is a form of writing practiced in Central Asia and Far East. It involves writing in clay tablets using wedge-shaped letters and baking those tablets. 
	 
	 Clay tablets, unlike stone, are more fragile. Therefore, even though the cuneiform script is cracked, it is difficult to transliterate the inscription due to wear and tear. 
	 
	 In \cite{cuneiform2006albayatti}, the authors propose a rule-based method to decode cuneiform. They convert Cuneiform imprints to "intensity curves" where a graph is drawn based on pixel intensities across a particular axis. Based on the shape of graph, the characters are classified into one of the known character sets.

    Another rule-based approach is used by \cite{character2015bogacz}, converts the characters into bezier splines and a rule based system is constructed to classify the control points of the splines.

    A rule tree is constructed based on the several features extracted using different image processing techniques, to classify the Cuneiform Letters in \cite{computerized2019aktas}

	 Some cuneiform tablets had broken into several pieces. \cite{gigamesh2010mara}, they use 3d CAD modeling to reconstruct the tablets and extract the text using photogrammetry.

    \begin{table}[hbt!]
    \centering
        \begin{tabularx}{\textwidth}{XXX}
        \toprule
        Paper & Method & Remarks \\
        \midrule
        \citeauthor{cuneiform2006albayatti} & Intensity Curves & Intensity curves are obtained by getting pixel intensities of the characters in a particular axis. Characters are classified by their unique intensity curves. \\
        
        \citeauthor{character2015bogacz} & Bezier Splines & Each stroke of the character is converted into a Bezier spline. Then they used a rule based system to classify those characters from the control points. \\
        
        \citeauthor{computerized2019aktas} & Multiple Image Processing methods and Rule based system & They process the individual Cuneiform characters using multiple image processing techniques. They use a rule tree to classify those characters. \\
        
        \citeauthor{gigamesh2010mara} & 3D modelling and CAD & This method uses 3D scan to construct the 3D model of the Tablet. Then they use CAD techniques to extract and identify the characters. \\
        \bottomrule
        \end{tabularx}
        \caption{Cuneiform Character Recognition}
        \label{}
    \end{table}

\begin{figure}
    \centering
    \includegraphics[scale=0.2]{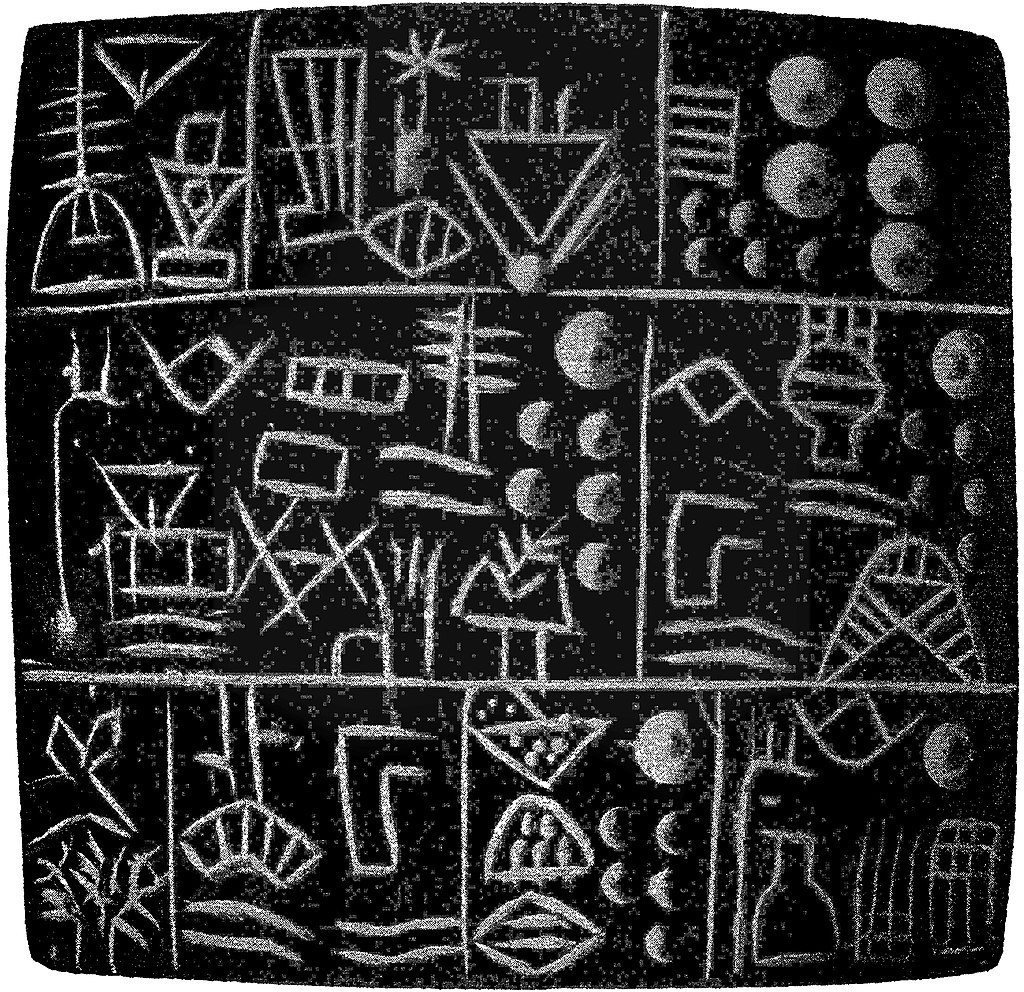}
    \caption{Imprint of Cuneiform Tablet}
    \label{fig:cuneiform}
\end{figure}

	 \subsubsection{Supervised Learning}
	 Supervised learning has proven to be a valuable tool for character recognition and transliteration, particularly when the script is already deciphered. In this section, we review several techniques that employ supervised learning for character recognition and transliteration in various scripts.

One approach is the use of template images to denoise inscriptions obtained from stone inscriptions, as demonstrated in \cite{multiscale2020preethi}. The authors perform image correlation on a noiseless template image of an inscription to enhance its clarity.

In \cite{feature2015soumya}, the authors utilize gradient and intensity-based filters to extract features from ancient Kannada language inscriptions. The features are then transliterated to modern Kannada symbols using fuzzy logic.

Scale Invariant Feature Transform is used to extract features and identify the decoration and the text areas in ancient manuscripts in \cite{detecting_garz_2010}

In \cite{feature2017suganya, ancient2013mahalakshmi}, complex transforms such as Shape and Hough transforms are used to extract features from epigraph images, and characters are classified using swarm optimization algorithms like Group Search Optimization and Firefly optimization. Also along with \cite{ancient2013mahalakshmi, digital_devi_2019} uses Median fuzzy filters to classify the characters. 

The use of Gabor filters for feature extraction in Old Latin and French texts is described in \cite{eighth2012rajakumar}. The authors evaluate word segmentation using several methods and show that Support Vector Machines (SVMs) perform well in this task.

\cite{recognition_alirezaee_2004} discusses a method, where set of invariant features are extracted from documents of middle age Persian origin. Then these features are used to classify the characters using Minimum Mean Distance Classsifiers and K-Nearest Neighbour Classifiers. 

SVMs are also used for character classification in cuneiform scripts in \cite{word2016homburg,automating2017bogacz}. In \cite{word2016homburg}, multiple SVM classifiers are used as an ensemble to classify cuneiform characters based on multiple extracted features. In \cite{automating2017bogacz}, SVMs are used to classify individual character images, which are then used in a Hidden Markov Model for further attribution.

In \cite{visual2016amato}, a Fisher vector is used for feature extraction, and a neural network is used for character classification. Chain code and Fourier transform features are used in \cite{combining2015chacko}, and two CNNs are utilized for character classification.

Various versions of Convolution Neural Networks (CNNs) are used for character recognition and classification in \cite{optical2018avadesh,data2018daggumati, digitalization2018sudarsan, novel2019giridhar, recognition2020gautam, convolution2020magrina, deepnetdevanagari2021narang, recognition2019wijerathna, improving_rajnish_2023, cnn_m_2022}. In \cite{data2018daggumati}, CNNs are used to compare Indus characters that are similar to Brahmi and Phoenician characters. The study demonstrates that Indus script visually resembles the Phoenician script more than the Brahmi script.

In \cite{recognition_ezhilarasi_2023}, the authors extract feature vectors from the image patches of rock carvings through Convolutional Neural Networks. The attention values of these feature vectors are calculated and these vectors are processed by LSTMs. The LSTM predicts the character in that patch of the image. 

However, the authors of \cite{one_liu_2022}, used one-shot approach to classify the characters and symbols. They train a Siamese Similarity Network to build a feature space which optimally places all the character classes for classification. They introduced a new loss called Soft Similarity Contrast Loss in addition to traditional Contrastive loss, to train the Siamese network. The resulting network classifies the characters even though it had seen very few examples of the character

The use of CNN-based auto-encoder architectures, such as U-Net and ResUNet, for transliterating ancient Japanese characters into modern Japanese is described in \cite{kuronet2019clanuwat}.

A semi-supervised method to label the characters of new scripts is introduced in \cite{semisupervised_vajda_2011}, where every character is clustered into buckets based on multiple levels of abstraction automatically. Then multiple human experts label those buckets and vote among themselves to decide whether a particular bucket is acceptable. This only consumes 0.5 times the time and effort taken in manually annotating all those labels. 

In \cite{hybrid2021kumar}, Capsule Networks are used instead of CNNs for character classification.

Finally, \cite{deep2017palaniappan} utilizes Google Le Net for classifying whether or not a character belongs to the Indus script. In \cite{deep2020dencker}, weakly supervised CNNs are employed, with individual CNNs for line alignment, line and word segmentation, character segmentation, and character recognition.

    \begin{longtable}{p{0.3\textwidth} p{0.3\textwidth} p{0.35\textwidth}}
        \caption{Supervised Character Classification methods} \\
        \toprule
        \textbf{Paper} & \textbf{Method} & \textbf{Remarks} \\
        \midrule
        \endfirsthead
        
        \multicolumn{3}{c}{\tablename\ \thetable{} -- Continued from previous page} \\
        \midrule
        \textbf{Paper} & \textbf{Method} & \textbf{Remarks} \\
        \midrule
        \endhead
        
        \midrule
        \multicolumn{3}{r}{Continued on next page...} \\
        \endfoot
        
        \bottomrule
        \endlastfoot
        \citeauthor{multiscale2020preethi} & Image Correlation &  Use a template image of a noiseless inscription and perform image correlation to denoise and classify the inscription obtained from a stone inscription \\
        \citeauthor{feature2015soumya} & Filtering & Gradient and Intensity filters are used to extract features and classify the ancient Kannada inscriptions\\
        \citeauthor{recognition_alirezaee_2004} & Invariant Moments & A set of invariant moments are used to classify Persian  Documents from Middle ages. \\
        \citeauthor{feature2017suganya} & Group Search Optimization and Firefly algorithm & Shape and Hough Transforms are used to extract features from the characters. GSO and Firefly algorithms are used to classify them.\\
        \citeauthor{ancient2013mahalakshmi, digital_devi_2019} & Particle Swarm Optimization &  Image of scripts are enhanced using Contourlet transform, denoised using fuzzy median filters and classified using Particle Swarm optimization\\
        \citeauthor{detecting_garz_2010} & SIFT & Scale invariant Feature transform is used to detect the text areas from decorative elements in manuscripts. \\
        \citeauthor{eighth2012rajakumar} & Support Vector Machines & Features are extracted from Old Latin and French texts which doesn't have inter-word spaces using Gabor Filter. The features extracted are then classified using Support Vector Machines\\
        \citeauthor{automating2017bogacz} & SVM and Hidden Markov Models & Characters of Cuneiform texts are segmented and ensemble of SVM and Hidden markov model is used to learn the sequence of the characters and classify them.\\
        \citeauthor{visual2016amato} & Fisher Vector and Neural Network & Fisher Vector of the individual characters are extracted as the features and they are classified using Neural Networks. \\
        \citeauthor{combining2015chacko} & Convolutional Neural Networks & Chain code and Fourier transform features extracted and used in two CNNs to classify characters\\
        \citeauthor{optical2018avadesh,data2018daggumati, digitalization2018sudarsan, novel2019giridhar, recognition2020gautam, convolution2020magrina, deepnetdevanagari2021narang, recognition2019wijerathna,improving_rajnish_2023, cnn_m_2022} & Various CNN models & Various CNN models such as AlexNet, ResNet, LeNet are used to classify the individual characters\\
        \citeauthor{one_liu_2022} & Siamese Network & One shot character Recognition using Siamese similarity network and Soft similarity Contrast Loss \\
        \citeauthor{data2018daggumati} & 	CNNs & Indus Script characters compared to Brahmi and Phoenician characters using CNNs, Indus script shown to be visually closer to Phoenician script \\
        \citeauthor{recognition_ezhilarasi_2023}& CNN-LSTM Network & The visual features are extracted using CNNs and those feature vectors are then processed by LSTMs to classify the characters. \\
        \citeauthor{kuronet2019clanuwat} & Autoencoder & Ancient Japanese characters transliterated to modern Japanese using CNN-based Auto-encoder architectures such as U Net and ResUnet \\
        \citeauthor{semisupervised_vajda_2011} & Semi-supervised Learning & Images are clustered based on different abstractions and human experts vote on to the labels assigned. \\
        \citeauthor{hybrid2021kumar} & Capsule Network & Characters classified using Capsule Network instead of CNNs\\
        \citeauthor{deep2017palaniappan} & LeNet & Characters classified whether they belonging to Indus script using Google Le Net \\
        \citeauthor{deep2020dencker} & Weakly supervised CNNs & Line alignment, segmentation of lines, words, characters and character recognition using individual CNNs. \\

    \end{longtable}

	 \subsection{Attribution}
	 Attribution refers to the task of assigning metadata to transliterated texts. This can be sequence/pattern prediction in case of undeciphered texts, place, missing text prediction, sequence prediction of broken tablets, place and time of origin prediction, etc. In some cases, the meaning of the epigraph can also be translated into modern language.
	 
	 Even though the characters are interpreted in the transliteration part, attribution enables the epigrapher to interpret the meaning of an inscription. 
	 
	 Traditionally, the epigrapher will use his domain knowledge to interpret an inscription, which is subjective and always debatable.
	 
	 In computational epigraphy, attribution is usually done by a system that understands the distribution of the lexicons in the particular script. This could be anything ranging from a frequency table to a deep learning NLP model.
	 
	 \subsubsection{Statistical Analysis of Indus Script}
	 As discussed above, Indus script is undeciphered, therefore several statistical analyses are done on those inscriptions attempting to decode it. \cite{frequency1988kak, statistical2010yadav, statistical2008yadav, segmentation2008yadav, statistical2019oakes, classification2011yadav, identifying2021daggumati} are research which attempted at deciphering Indus script using Statistical analysis. 
	 While most of them drew frequency tables for each Indus script character and tried to predict the next character based on probability, \cite{statistical2010yadav} used n-grams (considering an n-character window at a time) and tried to predict the characters. These studies revealed several underlying patterns within the Indus script.  For example, one of the results of \cite{frequency1988kak} is that the frequency tables of the Indus script and the Brahmi script (which is deciphered) are related. Similarly \cite{statistical2008yadav} showed several recurrent pattern of characters in Indus script. This shows that the Indus script is indeed a semantic script and has proper grammar, unlike pictorial scripts like Egyptian hieroglyphs. Similarly \cite{multiobjective2009das} used evolutionary algorithms to optimize KL divergence between distributions of Indus and Brahmi scripts and found the respective Brahmi characters that are most similar to Indus Characters in terms of semantics. 
	 
	 \cite{statistical2010snyder} designed a Bayesian Classifier to predict the probability of next character in Indus texts. 
	 
	 \cite{clustering2017yadav} attempts to group the characters in Indus Script into commonly used phenomes using K-Means clustering. \cite{computational2014yadav} uses same K-Means clustering algorithm to group Indus script into recurrent patterns. 
	 
	 \subsubsection{Reconstruction of Cuneiform Tablets}
	 We have mentioned above that Cuneiform script is written on clay tablets. Due to the nature of the medium, a lot of clay tablets are broken into several pieces. \cite{computerassisted2014collins, computational2017collins} discusses methods of reconstructing multiple pieces of cuneiform tablets in 3D CAD software by matching text on both the pieces. 
	 
	 In particular, \cite{computational2017collins} reconstructs the famous Atrahasis Cuneiform Tablet of which two pieces are present in Geneva and London. Due to large distance between the location of two pieces and fragility of the medium, there has not been any attempt to reconstruct both pieces into full tablet. This paper successfully confirmed that both pieces belong to same tablet and extracted the cuneiform text from it.

    A framework called Gigamesh is proposed by \cite{gigamesh2010mara}, where the broken cuneiform tablets are 3D scanned and transliterated using Giza++ program. 
	 
	In contrast, \cite{theoretical2013chng} discusses the use of stigmerty, a property that living organisms such as termites and bacterial colonies use to construct themselves into the most efficient shapes to reconstruct Cuneiform tablets. 
	 
	 \cite{photogrammetric2012lewis} uses photogrammetry,i.e., analyses the 2D images of Cuneiform tablet imprints and reconstructs them into full texts.  Naive Bayes, Maximum Entropy classifier is used in \cite{toward2012tyndall} to predict the probability that the text lines in one piece of a cuneiform tablet match the other piece. 

    \begin{table}[hbt!]
    \centering
        \begin{tabularx}{\textwidth}{XXX}
        \toprule
        Paper & Method & Remarks \\
        \midrule
        \citeauthor{computerassisted2014collins} & 3D CAD software reconstruction & Reconstructs multiple pieces of cuneiform tablets in 3D CAD software by matching text on both pieces.\\
        
        \citeauthor{computational2017collins} & 3D CAD software reconstruction & Reconstructs the Atrahasis Cuneiform Tablet by confirming that both pieces belong to the same tablet and extracting the cuneiform text from it.\\
        
        \citeauthor{gigamesh2010mara} & 3D scanning and transliteration & Proposes a framework called Gigamesh, where broken cuneiform tablets are 3D scanned and transliterated using Giza++ program.\\
        
        \citeauthor{theoretical2013chng} & Stigmerty-based reconstruction & Discusses the use of stigmerty to reconstruct Cuneiform tablets.\\
        
        \citeauthor{photogrammetric2012lewis} & Photogrammetry-based reconstruction & Uses photogrammetry to analyze 2D images of Cuneiform tablet imprints and reconstruct them into full texts.\\
        \citeauthor{toward2012tyndall} & Naive Bayes, Maximum Entropy classifier-based reconstruction & Predicts the probability that the text lines in one piece of a cuneiform tablet match the other piece using Naive Bayes, Maximum Entropy classifier.\\
        \bottomrule
        \end{tabularx}
        \caption{Cuneiform Character Recognition}
        \label{}
    \end{table}

	 \subsubsection{Predicting Metadata of the Inscriptions}

  \begin{figure}
      \centering
      \includegraphics[scale=0.1]{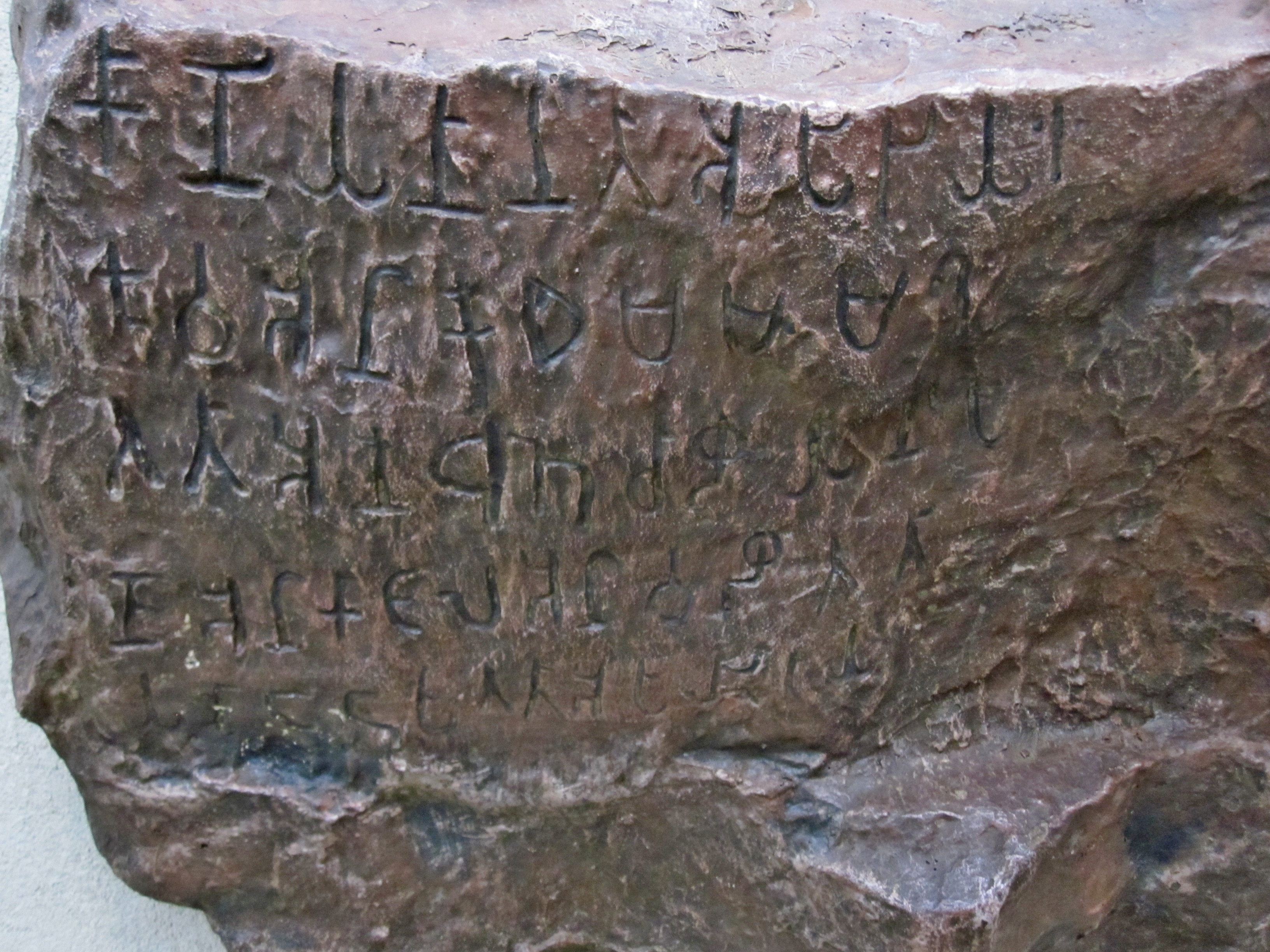}
      \caption{Imprint of Maangulam Tamil-Brahmi Inscription}
      \label{fig:Tamilbrahmi}
  \end{figure}

	 Epigraphs contain valuable information beyond what is explicitly written on them. For instance, Recchia et al. \cite{archaeology2016recchia} accurately predicted the place of origin and associated artifacts of Indus script inscriptions by using Semantic Latent Analysis, a commonly used method in computational linguistics.

    Based on the success of Handwriting recognition, the writers of Greek stone inscriptions are identified in \cite{handwriting_papaodysseus_2010}. The statistical features are extracted from the images are characters and based on those features, the writers are classified. Fractal Dimension Method is used to classify Ancient Arabic documents from Latin documents using \cite{characterization_zaghden_2011}

    The era on which the epigraphs are written in Kannada Language epigraphs are found using Support vector Machines according to \cite{svm_soumya_2011} and \cite{efficient_kumar_2014}. Here visual features are extracted and era of the epigraphs are classified by Support Vector Machine in \cite{svm_soumya_2011} and Transductive SVMs in \cite{efficient_kumar_2014}

\begin{figure}
    \centering
    \includegraphics[scale=0.25]{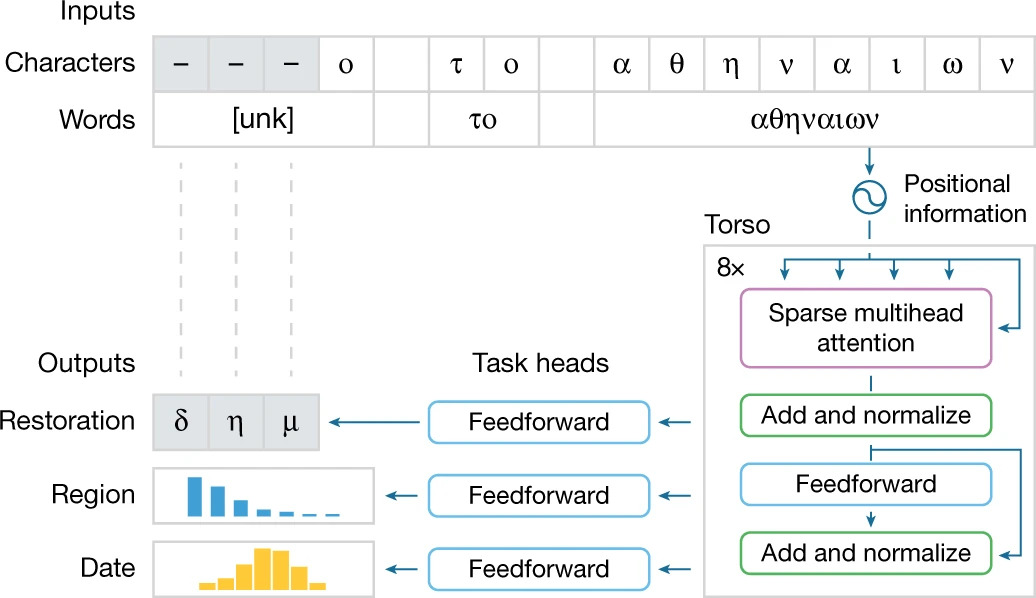}
    \caption{Architecture of model proposed by \cite{restoring2019assael} to restore and attribute Greek inscriptions}
    \label{fig:assael}
\end{figure}

    In a similar vein, Pabasara et al. and \citeauthor{period_surasinghe_2021} \cite{pabasaraPeriodPredictionSinhala2021, period_surasinghe_2021} predicted the period of origin of Sinhala epigraphs and extracted visual features of ancient Sinhalese using convolutional neural networks. Meanwhile, Vani et al. \cite{soft2020vani} used handcrafted features like Zernike Moments and HoG features to extract visual information from epigraphs. Similarly \cite{classification_soumya_2014} uses Zernike Moments and Normalized Central Moments extracted from segmented characters and uses Random Forest to Classify the characters. 

\begin{figure}
    \centering
    \includegraphics[scale=0.3]{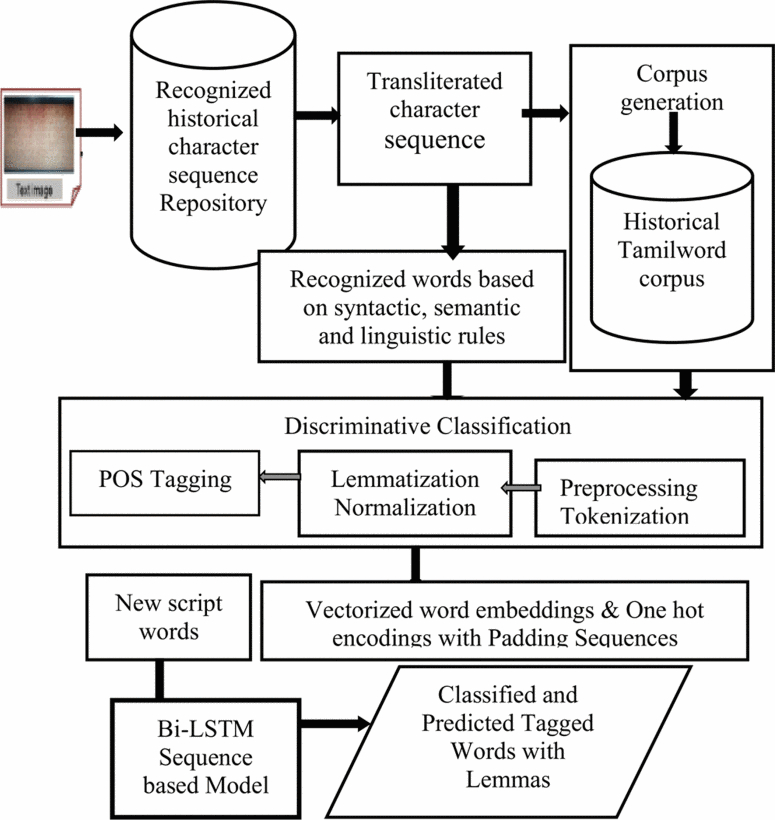}
    \caption{Architecture propsed by \cite{depicting_ezhilarasi_2021}, to do POS tagging and Lemmatization of Tamil Inscriptions}
    \label{fig:ezhilarasi}
\end{figure}
    Bi-directional LSTMs are used for Lemmatization and Part of Speech tagging in texts from ancient epigraphs in \cite{depicting_ezhilarasi_2021}
    
    Transliterated epigraphs often suffer from missing characters due to wear and tear, making them difficult to interpret. To address this problem, Assael et al. \cite{restoring2019assael, RestoringAttributingAncient} utilized Long Short-Term Memory networks and Transformers to predict missing words in Greek epigraphs and even predict the time and date of their origin. Saret et al. \cite{filling2021saret} employed LSTM and BERT to fill missing letters in Akkadian texts, while Fetaya \cite{restoration2020fetaya} used LSTM networks to predict missing characters in Babylonian script. Generative Adversarial Networks (GANs) along with Charbonnier Loss Function to denoise stone inscriptions and inpaint the missing segments of them in \cite{ancient_zhang_2021}
    
    In terms of translating epigraphs, Pagé-Perron et al. \cite{machine2017pagperron} used traditional NLP techniques such as stemmatization, morphological analysis, and POS tagging to translate Sumerian epigraphs to English, while Park \cite{ancient2020park} utilized Transformers to translate Ancient Korean to English.

\begin{figure}
    \centering
    \includegraphics[scale=0.15]{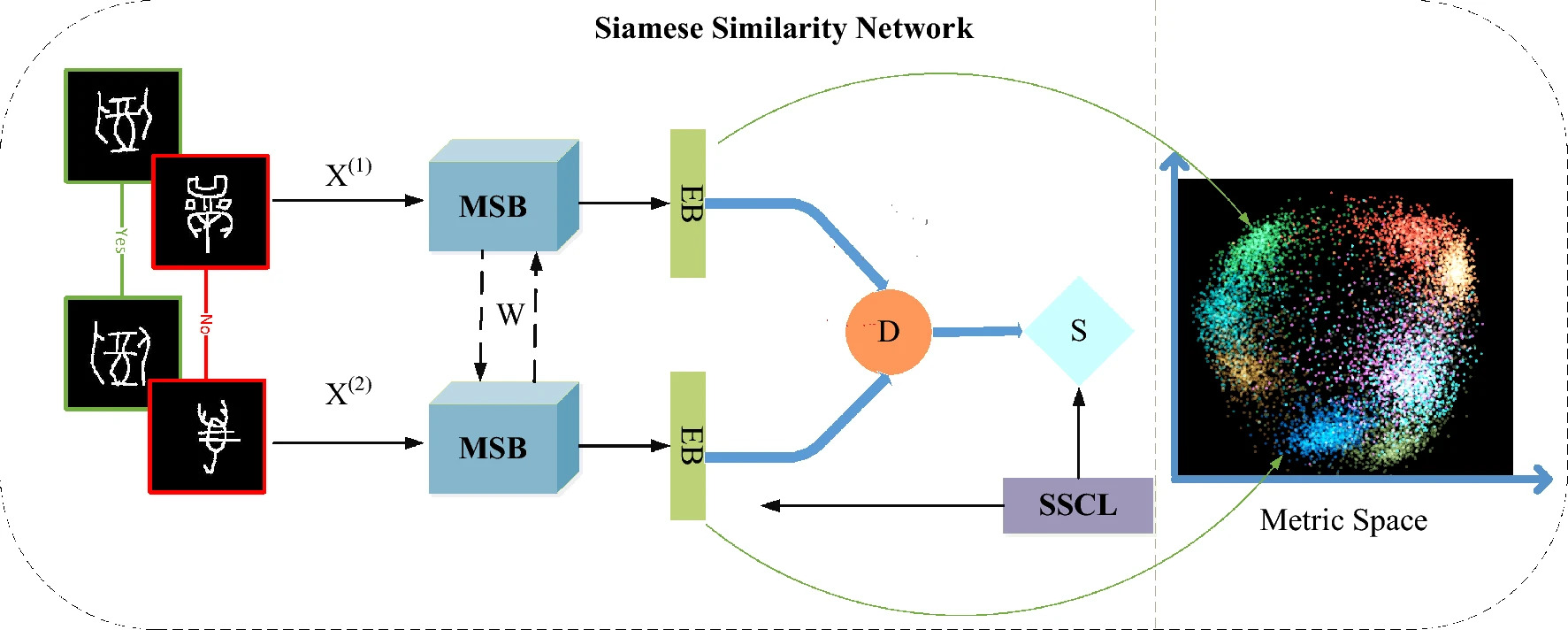}
    \caption{Architecture of Siamese Similarity Network proposed by \cite{one_liu_2022}. It mainly includes multi-scale fusion backbone structure (MSB), embedding structure (EB), fusion distance layer (D), similarity layer (S) and soft similarity contrast loss (SSCL)}
    \label{fig:siamese}
\end{figure}
    
    Finally, Choo et al. \cite{restoring2021choo} used Transformer architecture to restore and translate ancient Korean historical documents into modern English, and performed topic modeling to identify major themes in the documents.

\begin{longtable}{@{}p{3cm}p{4cm}p{7cm}@{}}
\caption{Attribution of Epigraphy} \\
\toprule
Paper & Method & Remarks \\
\midrule
\endfirsthead

\multicolumn{3}{c}%
{{\bfseries \tablename\ \thetable{} -- continued from previous page}} \\
\toprule
Paper & Method & Remarks \\
\midrule
\endhead

\bottomrule \multicolumn{3}{r}{{Continued on next page}} \\ \bottomrule
\endfoot

\bottomrule
\endlastfoot

\citeauthor{archaeology2016recchia} & Semantic Latent Analysis & Accurately predicts place of origin and associated artifacts using Indus script text\\
\citeauthor{pabasaraPeriodPredictionSinhala2021, period_surasinghe_2021} & Convolutional Neural Networks & Predicts period of origin of Sinhala epigraphs and extracts visual features using CNNs \\
\citeauthor{soft2020vani} & Handcrafted Features & Uses Zernike Moments and HoG features \\
\citeauthor{characterization_zaghden_2011}& Fractal Dimension Method & Fractal Dimension Method is used to classify Arabic Documents from Latin Documents. \\
\citeauthor{svm_soumya_2011, efficient_kumar_2014} & Support Vector Machines & SVMs are used to predict the periods of Tamil and Kannada Epigraphs \\ 
\citeauthor{classification_soumya_2014}& Random Forests & Zerinike and Normalized central moments are extracted from the images and Random Forests are used to classify them. \\
\citeauthor{handwriting_papaodysseus_2010} & Writer Classification by statistical methods & Finds the writer of the Greek inscription based on the statistical values of the segmented character. \\
\citeauthor{restoring2019assael, RestoringAttributingAncient} & LSTM, Transformer & Predicts missing characters in Greek epigraphs, predicts time and date of origin, provides explanations based on saliency maps \\
\citeauthor{depicting_ezhilarasi_2021} & Bidirectional LSTMs & Lemmatization and Part of speech tagging is done by Attention and Bi-directional LSTMs. \\
\citeauthor{filling2021saret} & LSTM, BERT & Fills missing letters in Akkadian texts \\
\citeauthor{restoration2020fetaya} & LSTM & Predicts missing characters in Babylonian script \\
\citeauthor{machine2017pagperron} & Traditional NLP Techniques & Translates Sumerian epigraphs to English \\
\citeauthor{ancient_zhang_2021} & GANs and charbonnier Loss & GANs and Charbonnier Loss function is used to denoise images of epigraphs and inpain the missing segments.\\
\citeauthor{ancient2020park} & Transformer & Translates Ancient Korean to English \\
\citeauthor{restoring2021choo} & Transformer, Topic Modeling & Restores ancient Korean historical documents, translates to modern English, identifies major themes \\

\end{longtable}

	\section{Conclusion}
	We have discussed in detail the use of computation to aid in Epigraphical and Archaeological discoveries. We have seen that computational methods are so beneficial to epigraphers as they have not only simplified the tasks of epigraphers but also opened new ways to interpret epigraphs. For example, we have seen several computational ways to decipher Indus script which has not been deciphered with traditional methods. We have also seen that it is important to attribute the epigraphs along with transliterations in order to obtain meaningful interpretations of epigraphs. 
	
	The advent of Machine Learning Results seem to provide huge breakthroughs in the field of Computational Archaeology, especially in the field of Epigraphy. We have seen several models have provided deciphered previously undecipherable inscriptions and also solved historical debates. Also, most models simplify the pipeline of interpreting those inscriptions which by conventional methods is a very complex process. Therefore we conclude that Computational methods are slowly becoming the default way of approaching epigraphy. 
	
	\section{References}
	    
		\bibliography{References}
	
\end{document}